# A Draft Memory Model on Spiking Neural Assemblies

João Ranhel, Joao H. Albuquerque, Bruno P. M. Azevedo, Nathalia M. Cunha, Pedro J. Ishimaru.
Dept. of Electronics and Systems
Universidade Federal de Pernambuco, Recife, Brazil
joao.ranhel @ ieee.org

*Abstract*—A draft memory (DM) model for neural networks with spike propagation delay (SNNwD) is described. Novelty in this approach are that the DM learns immediately, with stimuli presented once, without synaptic weight changes, and without external learning algorithm. Basal on this model is to trap spikes within neural loops. In order to construct the DM we developed two functional blocks, also described herein. The 'decoder' block receives input from a single spikes source and connect it to one among many outputs. The 'selector' block operates in the opposite direction, receiving many spikes sources and connecting one of them to a single output. We realized conceptual proofs by testing the DM in the prime numbers classifying task. This *activation-based* memory can be used as immediate and short-term memory.

## I. INTRODUCTION

To a certain degree, how nervous systems memorize and retrieve information remain unknown. Neuroscience have shed light in this area and it is common to create models in artificial neural networks (ANN) attempting to comprehend certain aspects of memorizing and recovering information. This article describes a fast-memory based on trapping spikes within neural loops, so this memory do not rely on changing synaptic weights. The network learns immediately, somehow like the extreme-learning paradigm. Erasing and changing information are easy operations. That is the reason why we call it a 'draft memory' (DM). We created the DM on a spiking neural network that takes into account the spike propagation Delay (SNNwD).

The term *memory* herein refers to the encoding, storage, and retrieval of learned information. Qualitatively, scientists use to classify human memories in two systems, generally referred to as *declarative* (conscious) and *non-declarative* (procedural) memories. They also classify memories according to the time over which they are effective, distinguishing the *Immediate* memory (*sensory* memory, lasting from fractions of a second to seconds), the *Working* or *short-term* memory (lasting from seconds to minutes), and the *Long-term* memory (lasting from days to years) [1],[2]. It is also possible to classify memory according to the properties of their underlying mechanisms; e.g. the changes in synaptic weights (*weight-based memory*), or the persistent activity (*activation-based memory*) [3]. Accordingly, this article describes an activation-based memory that can operate as both immediate and short-term memory, for declarative or non-declarative purposes.

In order to create the DM it was necessary to develop two blocks: the 'selector' and the 'decoder' block, both explained later. We worked on the Neural Assembly Computing (NAC) framework [4]; thus, these new blocks add functionalities to other NAC blocks. In fact, the DM uses decoders and selectors associated to other functional blocks described in previous works (see [5],[6],[7],[8]).

It is necessary to convert information (or datasets) to spikes in order to stimulate SNNwD. We tested the DM with a set of numbers associated to flags that classify them as prime or non-prime. We have codified numbers in regular binary and in Gray-code. Classification of prime numbers is a complicated task for weight-based ANNs. Prime numbers is a natural dataset with particular attributes: their distribution is chaotic, irregular, not smooth, and not patterned. Due to these characteristics, they do not fit well on traditional ANN classifiers. Functionally, DM operates as a prime identifier; but maybe it is better to consider DM as a memorizer instead of classifier. Somethings we simply memorize, such as times table, cities and people's names, etc. DM memorize sets of bits (any representation codified as spikes) and attributes associated to them. Therefore, DM can be seen as a topology for store and retrieve correlations on SNNwD; and we used the prime classifier just as a proof of concept.

The remainder of this paper is organized as follows: Section II contains the methodology used for creating and test the blocks and the DM. In Section III we describe the 'selector' and the 'decoder' functional blocks. Section IV holds a description of how to create a DM with blocks and other components. In Section V we present results, followed by a discussion about applications and further investigations. In Section VI we present conclusions concerning these neural circuits.

## II. METHODS

We created a Matlab code for simulating SNNwD using the NAC principles: neural assemblies execute logical functions, so they act as gates, samplers, and they can execute finite state machines, as well as to execute algorithms [4],[5],[6],[7],[8].

For simulations we used two neuron models: the Leaky Integrate-and-Fire neuron (LI&F), and the Simple Model (SM) [9], [10]. We first design a network topology for each block with predefined synaptic weights, according to some functional goal. Then, we adjust parameters in order to obtain best performance for some specific functional block [8]. Somehow, this method is similar to neural engineering [11], besides we use SNNwD and logical functions instead control theory concepts. Once the block is stably working, we inject noise in the network for testing its robustness.

## A. Neural Network

Rhythms are important components on NAC; thus, we start by creating pacemakers to generate rhythms that synchronize operations. Pacemakers are circuits that reverberate spikes inside the network structure.

According to the neuroscience literature, neurons operate in groups, in neural assemblies. Networks modeled this way operate with redundancy (e.g. [12]). Thus, one bit of information is represented not by a single spike, but by a group of spikes, in a *polychronized* pattern [13], using a sparse codification. However, when it is possible, we simulate NAC with a single neuron per assembly in order to increase the simulation speed. This is the case of DM, which requires only a spike and two neurons for trapping a bit (Fig. 4), therefore, the proof of concept for DM uses one spike per bit (Fig.5).

## B. Dataset and Spike Trains

Datasets have to be converted to spike trains for stimulating a SNNwD. Instead of choosing a popular dataset we opt for a natural one; thus, we converted a range of numbers (0 to 15) to spikes. On one hand, numbers are abstractions; hence, it is possible to represent them in several manners. On the other hand, spiking neurons can be firing or silent; thus, given this binary nature, we decided to represent numbers as binary spikes. We created a spike stream with numbers represented in binary (firing/not-firing) pattern, codified as regular binary code. Four spikes represent each number, with a digital value '1' meaning a present spike, and a value '0' meaning a silent neuron. Hence, "0000" means no neurons firing and "1111" means the four neurons firing synchronously.

As said before, rhythms are important, thus, we used five neural assemblies for creating a pacemaker. The assembly $P_1$ triggers $P_2$, which triggers $P_3$, which triggers $P_4$, which triggers $P_5$, which triggers $P_1$. This loop remains firing indeterminately. We called phases ($\varphi1$, $\varphi2$, $\varphi3$, $\varphi4$, and $\varphi5$) the time each of these neurons fire. Remember that spikes takes time to propagate in SNNwD; thus, we used a delay of 20 *ms* among the pacemaker assemblies. It means that the pacemaker cycle repeats each 100 *ms*. Every time $P_1$ fires ($\varphi1$) we introduce a new binary number associated to a set of control spikes (M, R, E, explained later), and two attribute bits informing whether the number is prime ($\Pi$) or not ($n\Pi$). In tests, numbers were stored in sequence, but the DM can store and retrieve spikes in any sequence. M stands for 'memorizing', E for 'erasing', R for 'retrieving', while $n\Pi$ informs that the number is non-prime, and $\Pi$ informs that the number is prime.

In order to show that the memory can record any type of representation, we created a second spike stream with the same dataset codified in the Gray code. Again, a spike represents a '1' and a silent neuron represents a '0' in the Gray code. In the tests, numbers are also stored in sequence, but due to the different codification in Gray, numbers are stored in or retrieved from different memory cells, compared to the binary code.

In order to record or retrieve values and correlations in DM, we must tell in the spike stream what we want to do. Thus, we used three other input control spikes in the stream that inform if the number and the attributes must be memorized (the input M), if the memory cell must be erased (the input E), or if the attribute of certain memory cell must be retrieved (the input R).

As an example, if we want to store the number '5' in the DM, the spike stream appears as in the Table I (shown for both, binary and Gray code). The number '5' appears in $D_3$, $D_2$, $D_1$, and $D_0$. At the same time ($\varphi_1$), the M input is firing; commanding the DM to memorize the attributes, while the input $\Pi$ is firing, indicating that the number is prime.

TABLE I. SAMPLES OF THE #5 ON THE SPIKE STREAM

| | | (Spike Stream) – #5 as input to the network at $\varphi1$ period. | | | | | | | | |
|---|---|---|---|---|---|---|---|---|---|---|
| Dec | | $D_3$ | $D_2$ | $D_1$ | $D_0$ | M | R | E | $n\Pi$ | $\Pi$ |
| 5 | Bin | 0 | 1 | 0 | 1 | 1 | 0 | 0 | 0 | 1 |
| 5 | Bin | 0 | 1 | 0 | 1 | 0 | 1 | 0 | ? | ? |
| 5 | Gray | 0 | 1 | 1 | 1 | 1 | 0 | 0 | 0 | 1 |
| 5 | Gray | 0 | 1 | 1 | 1 | 0 | 1 | 0 | ? | ? |

Table I also shows how to retrieve attributes from the DM. The spike stream presents a set corresponding to the number ($D_3$…$D_0$) and, at the same time ($\varphi_1$), the R input fires, keeping M silent, commanding the DM to retrieve data. After a delay, the attributes appear in a 'reading gate', as explained later. As said, to store, to erase, and to retrieve data can be done in any order, all controlled in the stream by these input control spikes.

## C. Noise and Results

For both neuron models (LI&F and SM) we use the Matlab normal Gaussian function *randn()* for injecting noise in the network, just as described in [7]. The program inject noise at each simulation iteration and the temporal resolution of the simulation is 1 *ms*; thus, the noise frequency may be as high as 1 KHz. Noise is a percentage of the synaptic current directly injected to each synapse. As we increase noise, we qualitatively verify if the network still performs the designed function. We obtain the results visually in a raster plot, as shown later, but the program allows us to plot the membrane potential and synaptic weights, eventually used for adjusting the overall network synaptic weights and topologies.

## III. CREATING THE FUNCTIONAL BLOCKS

In NAC we use to start a network design by choosing basal functions on information processing, then we design topologies that perform such processing. We call them functional blocks. As examples, we consider a principle the network to select one among several stimuli sources. Thus, the 'selector' is capable of selecting one among many spike sources to propagate along the network. We also consider a principle the capability of choosing a specific memory place for storing or retrieving spikes among several possible locations, this is what the 'decoder' block do. Animals have large numbers of sensors that convert some type of energy to spikes. In many situations, such information sources compete to send spikes to a single region. The opposite is valid, and sometimes a punctual source generate spikes that take one path instead of many others within the neural network. These examples shows that there must exist neural circuits that distribute and others that concentrate spikes streams, which are complimentary functions executed by the mentioned blocks.

It is possible to implement the same function in several manners with different neural topologies. Next, we present the 'selector' and then the 'decoder' for SNNwD. Later, we combined them with gates [7] and bistable loops [5] in order to design a memory block with *m* cells that holds *n* spikes/bits.

*A. The Selector / Function Generator Block*

As said, we call 'selector' a neural circuit that receives spikes from several input sources but selects only one and connects it to a unique output, ignoring all the others.

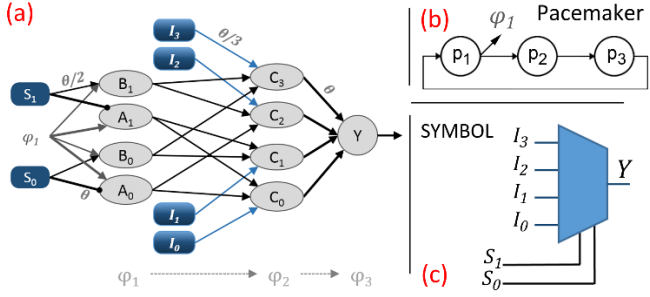

Fig. 1. Four-Input Selector. (a) This topology receives 4 spike sources from inputs $I_3$, $I_2$, $I_1$, $I_0$. Neurons $S_0$ and $S_1$ control which input is connected to the output Y. This topology acts as a selector but also as a function generator $Y=f(S_1,S_0)$. (b) A loop (pacemaker) generates the phases $\varphi_1$, $\varphi_2$, and $\varphi_3$. (c) The symbol used for this block (phases omitted in the symbol).

Fig. 1 shows a topology that performs this function. It receives four sources ($I_0$, $I_1$, $I_2$, and $I_3$) but only one at the time is able to propagate spikes to the neuron Y. In order to control which source is connected to Y the block requires two control spikes $S_1$, $S_0$. The pacemaker neuron $P_1$ ($\varphi_1$) and the control spikes ($S_0$, $S_1$) are connected to the neurons $A_0$, $B_0$, $A_1$, and $B_1$.

We explain now how we choose the synaptic weights when designing a topology. When the sum of all excitatory inputs is 'enough', it causes the neuron to fire. Note that we do not say 'threshold', because the simple model neuron does not deal with threshold [9]. We denote $\theta$ the excitatory post-synaptic potential (EPSP) that triggers a spike.

For instance, $P_1$ ($\varphi_1$) is connected to $A_0$ and $A_1$ by synaptic weights that cause $\theta$ on both. Thus, every time $\varphi_1$ occurs these neurons would fire. On the other hand, $P_1$ ($\varphi_1$) is connected to $B_0$ and $B_1$ by weights that causes $\theta/2$; hence, $\varphi_1$ cannot trigger these neurons alone, although every time $\varphi_1$ occurs $B_0$ and $B_1$ receive half of the EPSP they need to fire.

When $\varphi_1$ occurs, if $S_0$ and $S_1$ are both silent (not firing), $A_0$ and $A_1$ receive $\varphi_1$ excitation but no inhibitions; neither from $S_0$ nor from $S_1$. Therefore, $A_0$ and $A_1$ fires due to excitation coming from $P_1$. These neurons, as well as the input neuron $I_0$, project their axons to $C_0$. Each of these connections have a synaptic weight that causes $\theta/3$ in $C_0$. Therefore, the coincidence of these three spikes can trigger $C_0$, and it occurs only when $S_0$='0', $S_1$='0' (then $A_0=A_1$='1'), and the input $I_0$='1'. Note that, even if the input $I_0$ is not active, the neuron $C_0$ is the only one selected for propagating an input signal. The axons from all neurons $C_j$ are connected to the output neuron Y by a synaptic weight that causes $\theta$ on it; hence, any spike from $C_j$ neurons can trigger Y, this is equivalent to an OR function.

The propagation delay from $A_0$, $B_0$, $A_1$, and $B_1$ to the neurons $C_j$ is $\Delta t$ (*ms*), which is equal to the pacemaker time base. Only one $C_j$ neuron is able to fire at the time $\varphi_2$. Yet, the propagation delay from $C_j$ to Y is also $\Delta t$; hence, the Y output neuron fires synchronized to $\varphi_3$.

Now let us suppose $S_0$='0' and $S_1$='1' at $\varphi_1$. The spike from $\varphi_1$ trigger $A_0$ and tries to trigger $A_1$; however, the spike from $S_1$ inibits $A_1$. At the same time, spikes from $\varphi_1$ AND $S_1$ coincide in $B_1$; thus, this neuron fires. The picture is that $B_1$ and $A_0$ are both firing, and as their axons are connected only in the $C_2$ neuron, the unique neuron capable of propagating a spike is $C_2$, that receives the input from $I_2$.

The reasoning is the same for all combinations of $S_0$ and $S_1$. Thus, it is possible to summarize the functionality of this topology this way: at the time $\varphi_1$, the selection neurons ($S_0$ and $S_1$) control which neuron pair ($A_i$, $B_i$) fires. At the phase $\varphi_2$ a single $C_j$ neuron fire, and at the time $\varphi_3$ the Y neuron propagate a spike coming from one of the inputs $I_j$, if there is a spike in the $I_j$ input. We can formalize the Y output as:

$$Y \leftarrow \bar{S}_1 \wedge \bar{S}_0 \wedge I_0 \vee \bar{S}_1 \wedge S_0 \wedge I_1 \vee S_1 \wedge \bar{S}_0 \wedge I_2 \vee S_1 \wedge S_0 \wedge I_3 \quad (1)$$

This formula is read this way: *Y* is caused by (*not* $S_1$ AND *not* $S_0$ AND $I_0$) OR (*not* $S_1$ AND $S_0$ AND $I_1$) OR ($S_1$ AND *not* $S_0$ AND $I_2$) OR ($S_1$ AND $S_0$ AND $I_3$). The formula denotes ephemeral phenomena that happen only during $\varphi_3$, reason why it is not a Boolean equation, but a formula that denotes a causal relation. After some algebra we deduce that $Y=f(S_1,S_0)$; thus, by connecting the control inputs to '1' ($I_j$=1, e.g. $I_j$=$\varphi_2$), or not connecting them at all ($I_j$='0'), we obtain a Boolean function generator. This is possible because the $A_i$ and $B_i$ neurons predict all products (ANDs) that $S_1$ and $S_0$ can generate, while the Y neuron perform an OR function of all terms it receives.

It is intuitive to infer that a selector with $\Omega$ control inputs is capable of selecting one out of $2^\Omega$ inputs. However, it is not possible to follow such general rule when constructing selectors in NAC. As shown in [8], there are limits for executing AND functions in spiking neurons. For three control inputs ($S_1$, $S_1$, $S_0$) and eight spike sources, an AND at the $C_j$ would have four inputs. This is feasible because $\theta/4$ per input makes a reliable AND even in noisy situations. However, more inputs means that each inputs contribute less for EPSPs. For *n* inputs, each input contributes with a $\theta/n$ EPSP; thus, for the same level, noise becomes proportionally more relevant for each input.

*B. Decoder / Distributor block*

We call 'decoder' a topology that receives a single input and then distributes the input spikes to one among many outputs.

Fig. 2 shows a topology capable of performing this function. This neural circuit receives only one input spike source I, which is connected to four outputs ($Y_0$, $Y_1$, $Y_2$, $Y_3$). Similar to the previous block, in order to control which output receives the input spikes, the block requires two control spikes $S_1$, and $S_0$. The pacemaker neuron $P_1$ ($\varphi_1$) and the control spikes ($S_0$, $S_1$) are connected to the input selector neurons $A_0$, $B_0$, $A_1$, and $B_1$, which operates exactly as explained for the selector block.

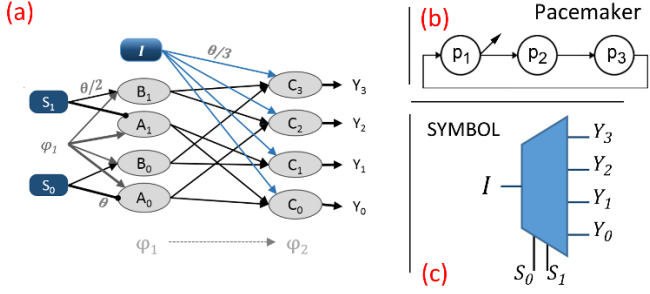

Fig. 2. Four-Output decoder. (a) Based on spikes from $S_1$ and $S_0$, this topology can connect the input I with only one of the four outputs $Y_0$, $Y_1$, $Y_2$, or $Y_3$. (b) The block is syncronized to the pacemarker circuit. (c) The symbol used for this block (phases omitted in the symbol).

This block is easier to understand because, instead of four inputs, this topology has only one input I connected directly to the neurons $C_0$, $C_1$, $C_2$ and $C_3$ on the output layer. As seen before, the layer with the pairs $A_0$, $B_0$, and $A_1$, $B_1$ fires according to the combination of $S_0$ and $S_1$. At the time $\varphi_2$, the $A_0$, $B_0$, $A_1$, and $B_1$ neurons enables only one $C_j$ neuron; the neuron that receives an EPSP of $2\theta/3$. The third input ($\theta/3$) necessary for triggering $C_j$ comes from the single input I. The output $C_j$ fires when I is '1', otherwise it remains silent even though $S_0$ and $S_1$ enable the $C_j$ output. The output of this block happens at the time $\varphi_2$.

Similar to the previous one, with $\Omega$ control inputs this block would be capable of connecting the input I to one out of $2^\Omega$ outputs. However, the same limitation concerning AND functions applies here. A good practice is to keep the number of input controls ($\Omega$) low, but arrange decoder blocks in parallel, as can be further seen in the DM project.

Four independent equations describe the $Y_j$ output:

$$Y_0 \leftarrow \bar{S}_1 \wedge \bar{S}_0 \wedge I; \qquad (2)$$

$$Y_1 \leftarrow \bar{S}_1 \wedge S_0 \wedge I; \qquad (3)$$

$$Y_2 \leftarrow S_1 \wedge \bar{S}_0 \wedge I; \qquad (4)$$

$$Y_3 \leftarrow S_1 \wedge S_0 \wedge I \qquad (5)$$

These formulas are read as the former, e.g. $Y_3$ is caused by $S_1$ AND $S_0$ AND I.

This block has two functions clearly defined. Firstly, it distributes information coming from I to one out of $2^\Omega$ output paths, so it works as a switch that distributes information among different data paths. Besides, because the neurons $A_i$, $B_i$ predict all products of $S_1$ and $S_0$, this block functions as a 'decoder' of $S_1$ and $S_0$ as we set the input I='1'. This is accomplished by connecting I to the $P_2$ ($\varphi_2$) neuron of the pacemaker.

## IV. A DRAFT MEMORY MODEL

As said, this article describes a memory that do not rely on synaptic weight changes. An option for creating a memory in SNNwD is to trap spikes within neural loops. We call 'bistable loop' the topology that catches a spike, and we call 'memory cell' a set of parallel bistable loops that retains information. We expecte that memories store/retrieve more than one set of bits (e.g. computer memories deal with several bytes); thus, we shall first describe the topology for selecting the memory cells, then we describe the cells, and then the memory operation.

### A. Memory Diagram

Fig. 3 shows the memory diagram divided in two operational unities. Fig.3a shows two decoder blocks combined to a layer of neurons resulting in a selection topology capable of 'addressing' sixteen memory cells. In the selection layer, the $Sel_k$ neuron generates a spike for selecting the *k* memory cell for storing, for retrieving, and for erasing spikes.

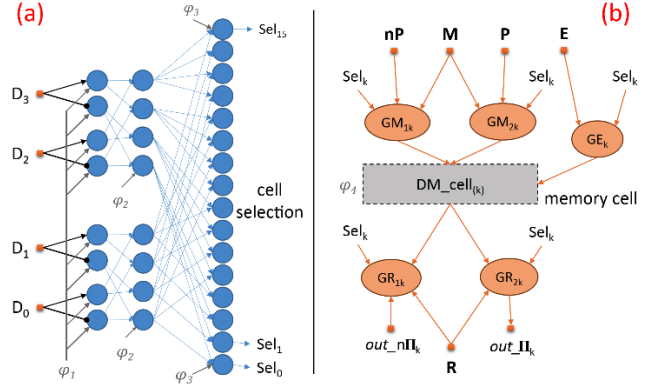

Fig. 3. Memory Diagram. (a) The memory cell selection uses $D_3$, $D_2$, $D_1$, and $D_0$ for decoding one among $2^4$ memory cells. (b) The input gates on each cell: the $GM_k$ gates activate for memorizing data, the $GR_k$ activate for retrieving data, and $GE_k$ gates activate for erasing data in the *k* memory cell.

Considering datasets modeled as *objects* having *attributes*, the DM is designed with objects (the numbers) addressing a memory location and attributes (the classes prime/non-prime) filling out the contents of the memory cell. It is possible to design memories with other configurations, but it is out of the scope of this article to provide an extensive discussion on these configurations.

Operations on the DM start when the stream has the number ($D_3$, $D_2$, $D_1$, and $D_0$) and the spikes M, E, R, nΠ, Π active at $\varphi_1$. It takes two additional phases for the $Sel_k$ neurons to address a memory cell (active at $\varphi_3$); thus, there are delay lines in the M, E, R, nΠ, Π inputs (not shown in Fig. 3). Delays exist for correct operation of these signals at $\varphi_4$ in the cell gates, the next layer the DM topology.

On *memorizing* operations, the object ($D_3$, $D_2$, $D_1$, and $D_0$) triggers a single $Sel_k$ neuron at $\varphi_3$, which is one condition ($\theta/3$) for the input gates $GM_{1k}$ and $GM_{2k}$ to be open (Fig.3b). Data storage occurs when M is firing, the second condition ($\theta/3$) for opening the input gates on the $DM\_cell_k$ cell. The third condition is the spikes from the attributes (the classes nΠ, and Π).

On *retrieving* operations, the object ($D_3$, $D_2$, $D_1$, and $D_0$) and the R input fire together in the spike stream. The $Sel_k$ spike happens at $\varphi_3$ and is one condition ($\theta/3$) for enabling the reading gates $GR_{1k}$ and $GR_{2k}$ (Fig.3b). Associated to the R input (the second condition $\theta/3$) they enable the reading gates, which copy the spikes trapped within the memory cell. These spikes can then propagate throughout the output ports 'out_nΠ$_k$' and 'out_Π$_k$'.

The *erasing* operation requires the object address a memory cell and a spike in the E neuron. We designed this memory for erasing contents on individual memory cells. An empty memory cell has no spikes trapped within the bistable loop, because the basal neurons on the cell is not reverberating, however, an active

(not erased) memory cell has their kernel neurons firing even if no spikes is trapped within their bistable loop, as explained next.

## B. Memory Cells

Each memory cell can store and retrieve *N* bits, although we have used only two bits in the present proof tests. A memory cell in DM has a basal reverberating kernel to which is possible to aggregate any number of bits, as shown in Fig. 4. Two neurons (Ka and Kb) form the kernel in the memory cell.

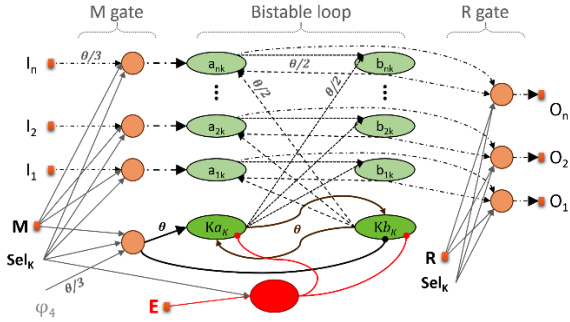

Fig. 4. Memory Cell. Spikes pass through the M gate (when memorizing) and remain 'trapped' into a bistable loop. They can then be retrieved through the R gate. Please, refer to the text for detailed explanation.

When a 'memorizing' operation is triggered ($\varphi_4$) the Ka neuron fires and the Kb neuron is inhibited. Then, Ka triggers Kb after $\Delta t$ ms, and Kb triggers Ka back after another $\Delta t$. They are connected by a synaptic weight that causes $\theta$ one another, so the loop remains operating regardless of a bit '1' or '0' is stored in each independent bistable loop. Only an *erasing* operation can dismantle this kernel, because this operation inhibits Ka and Kb simultaneously, and it forces both to become silent.

At the same phase $\varphi_4$, the M gate opens for all bits; thus, the firing input neurons ($I_1 \dots I_n$) trigger their correspondent pair in the left column of the reverberating loop (neurons $a_{1k}$ to $a_{nk}$). Neurons at the left column fire at the same phase as Ka in the kernel. After $\Delta t$, spikes from $a_{ik}$ neurons reach neurons on the right column of the bistable loop ($b_{1k}$ to $b_{nk}$). Neurons from the left column can cause only $\theta/2$ on their pairs in the right column; however, the Ka neuron is connected to all neurons in the right column, and it can cause the complimentary $\theta/2$ on all of them. Thus, Ka provides half of EPSP necessary for all neurons at right, and each firing neuron at left can trigger the correspondent $b_{jk}$ neuron at right. These connections are reciprocal; hence, Kb provides half of EPSP necessary for all neurons at left, and each firing neuron at the right column can trigger the correspondent $a_{jk}$ neuron at the left column.

A spike that passes through the M gate (at $\varphi_4$) triggers a single neuron $a_{jk}$ (neuron *a* of a bit *j* in the cell *k*) at phase $\varphi_5$. If a 5-phases pacemaker controls the rhythm, the neurons $a_{jk}$ first fire synchronized to $\varphi_1$, because after $\varphi_5$ the phase $\varphi_1$ repeats. In this sense, a spike remain trapped in $a_{jk}$ ($\varphi_1$) and $b_{jk}$ ($\varphi_2$), then $a_{jk}$ ($\varphi_3$) and $b_{jk}$ ($\varphi_4$), and then a phase inversion occurs for the next spike reverberation, with $a_{jk}$ ($\varphi_5$) and $b_{jk}$ ($\varphi_1$). Remember that a retrieving operation starts at $\varphi_1$, the memory cell selection is ready at $\varphi_4$, but as described above, the trapped spike can fire in $a_{jk}$ or in $b_{jk}$. That is why axons from $a_{jk}$ and $b_{jk}$ are both connected to the R gate. Certainly, an even number of phases on the pacemaker suppresses the phase shifts. In this case, neurons in the left column fire at odd phases ($\varphi_1$, $\varphi_3$, etc.) and neurons in the right column fire at even phases ($\varphi_2$, $\varphi_4$, etc.).

Note that retrieving data from memory cells is a non-destructive operation, so spikes remain preserved within the bistable loops. We designed DM with sixteen memory cells, each cell with two bistable loops that stores two attributes.

## V. RESULTS AND DISCUSSIONS

The DM can retain any information codified as spikes. As said, we used numbers and two classes (primes / non-primes) in order to test the DM. It is important to clarify that DM do not learn the concept of prime numbers. It holds and retrieves any correlations without generalization, with semantic meaning that users attribute to data stored on it.

Fig. 5 shows the results for the tests in two raster plots. At the top, Fig. 5a shows the storing and retrieving of regular binary numbers, while at the bottom, Fig. 5b shows the raster for similar operations in Gray code.

At the bottom of each raster there is a shaded area showing the spike input streams. As said, in order to easily understanding the plots, numbers were stored in sequence, although they were randomly retrieved. Below the plots, numbers indicate when they appear in the stream. In each raster, from bottom up, the first spike on the shaded stream zone is $\varphi_1$; then, appear $\varphi_2$, $\varphi_3$, $\varphi_4$, and $\varphi_5$, creating a rhythmic stairway-like pattern.

As said, $\varphi_1$ marks the time each number appears in the spike stream along with the M, R, E, nΠ and Π spikes. Remember that it takes time (from $\varphi_1$ to $\varphi_4$) for memorizing an information. Then, note in the raster plots that two reverberating spikes (denoted $M_0$) start firing near the second occurrence of $\varphi_1$. These are $Ka_0$ and $Kb_0$, the kernel neurons of the memory cell 0.

At the second occurrence of $\varphi_1$, the DM receives the command for memorizing the number 1; hence, a line with two reverberating neurons appears above $M_0$, formed after $Ka_1$ and $Kb_1$ starts firing. These are the kernel neurons of cell 1.

In the sequence, at the third occurrence of $\varphi_1$, the number 2 is stored, and this is a prime number. In Fig. 5a, note that the pattern for the memory cell 2 (the second spike set above $M_0$) is different from the formers. This is because the kernel $K_{a2}$ and $K_{b2}$ is firing associated to the spike Π in cell 2. Arrows indicate the cells of prime numbers (2, 3, 5, 7, 11, and 13). They fire different patterns compared to the non-prime memory cells.

In Fig. 5, the data retrieving starts at the middle of the plots. The numbers below the spike stream show the order they were 'consulted'. In order to better visualize the answers, the output R gates in all memory cells are connected to a unique set of neurons called 'Π*ans*'. The connection of the output of all memory cells to a single answer neuron resembles the connection of the input sources connected to a single output neuron on the selector block. Π*ans* fires a burst whenever the spike stream receives a prime number along with the R spike, as shown on the top of Fig. 5a and Fig. 5b. As said, the spike stream introduces the number and a spike in R synchronized to $\varphi_1$. Fig. 5a shows the retrieved numbers in the following order: 0, 8, 4; and none of them triggers the *Πans* neurons.

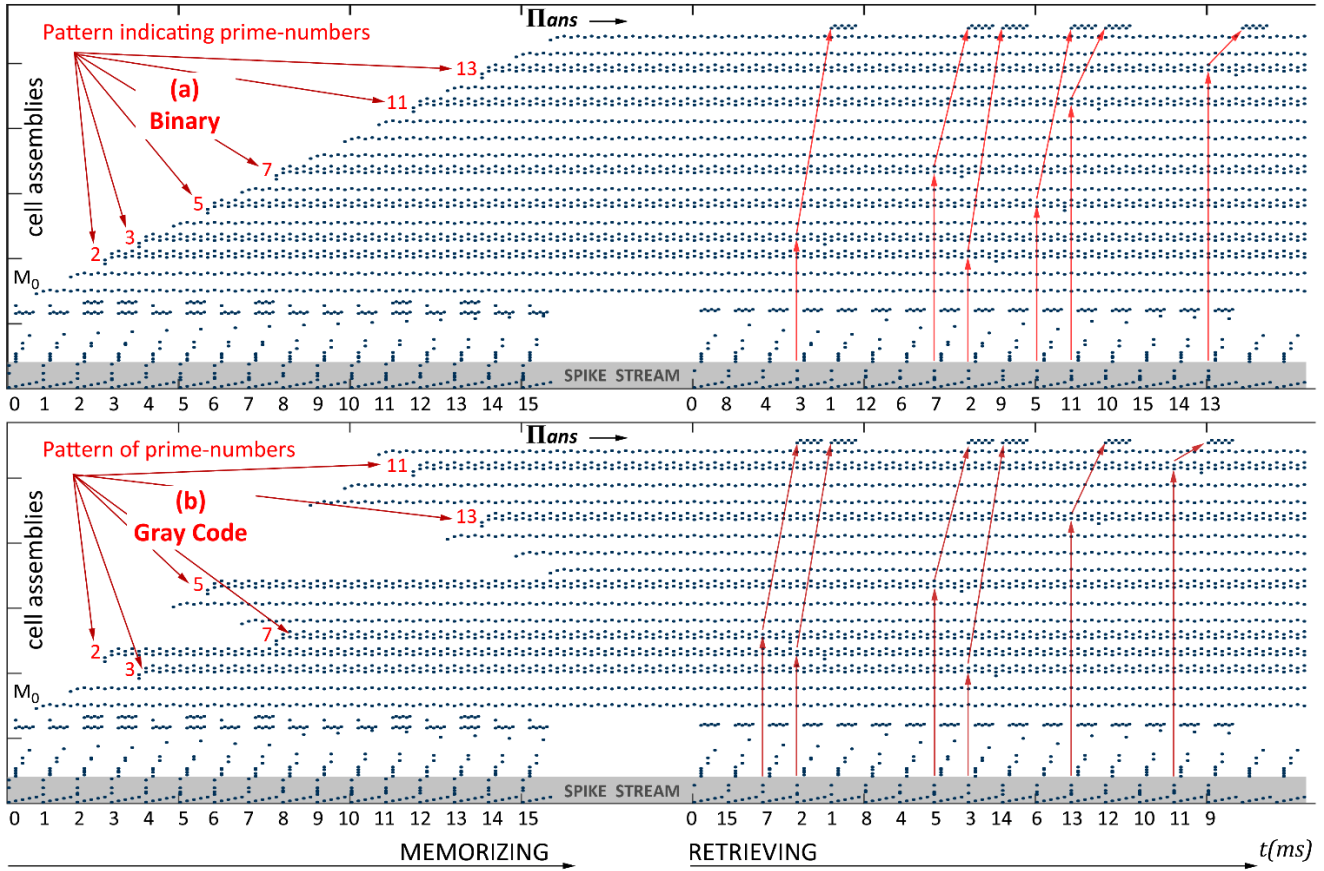

Fig. 5. Raster plot of the simulations. Memorization starts at 5 ms in both rasters, with numbers stored in sequence. Numbers and attributes are retrieved next, in random order. The order is denoted below the graphics. (a) Binary numbers. (b) Binary numbers in Gray code.

At this point, the spike stream introduces the number 3 and, after a delay, Π*ans* fires a burst. Then, the stream introduces the numbers 1, 12, and 6; and none of them triggers Π*ans*. After introducing the numbers 7 and 2, two *Π*ans bursts appear in sequence, followed by a silence as the number 9 is introduced, and then two *Π*ans bursts appears again, for the number 5, and 11. *Π*ans fires again only after introducing the number 13.

In Fig. 5, arrows shows the reading of prime numbers. They point from the moment a number appear in the stream, to the memory cell that stored its attribute, to the moment the answers appear in the raster plot, as a *Π*ans burst.

The DM works the same way for numbers codified on Gray code (Fig. 5b), although the raster seems different. The Gray code represents numbers by changing only one bit from a representation to another. For instance, the number 0="0000", 1="0001", 2="0011", and the number 3="0010". Only one binary digit is flipped in successive representations.

In Gray code, the number 2 is equal to the representation of the number 3 in binary, and vice-versa. The change in the object representation (the numbers in Gray) causes an exchange in the memory position used for storing them. Remember that we used binary combination in the decoder block for selecting a single output. Moreover, we used such outputs for addressing the memory cells. Thus, we caused an exchange on the memory position of a Gray number compared to the memory position of the same binary number. Nevertheless, each memory cell maintain the correlation with the attribute (prime or non-prime), because such attributes are independent of the memory cell used for store and retrieve them.

Note that we created a different retrieve sequence (Fig. 5b) for Gray code numbers and all the answers are correct. The arrows point out that, when the input stream presents a number for retrieving, the reading mechanism in the memory cells points to the correct firing pattern of that Gray number, so the Π*ans* spike burst appears only for the prime numbers.

*A. Discussions*

Why to create a immediate draft memory that store spikes instead of changing synaptic weights? Firstly, organic neural networks probably have a type of memory that stores information presented only once. The DM is a model for this kind of *activation-based* memory[3]. The connectionist paradigm, based on weight changes, in general have datasets presented more than once for the network. However, in natural SNNs, the stimuli may not appear repeatedly until the network fully change the synaptic weights. In this direction, we created the DM that stabilizes data as soon the network receives stimuli.

Secondly, spike-time dependent plasticity (STDP) is a well-stablished method used in SNN for implementing the 'temporal Hebbian rule' of synaptic weight changes. STDP is widely utilized in models of plasticity and learning. In STDP, a synapse is strengthened when a pre-synaptic spike reaches the synapse and then the post-synaptic neuron fires. On the opposite direction, a synapse is weakened when a neuron fires and later a spike reaches such synapse, coming from a pre-synaptic neuron. Temporal relations matter for STDP. Yet, STDP is a *protocol* for investigating long-time potentiation (LTP) and long-time depression (LTD) [14], [15]. The STDP protocol requires repeated pairing spikes, for 50-100 times, at a fixed frequency. Depending on the neural firing frequency, the protocol can last minutes to be complete. Another reason for creating the DM is to keep data stable for applying the STDP protocol without needing to excite the network with the same dataset several times. We intend to proceed with investigations in this direction.

The two functional blocks presented in this article open new perspectives on creating *datapaths* on NAC approach. Datapath and data storage can improve the manipulation of the spike streams within the SNNwD. Next steps in this direction is to use the recently created blocks associated to finite-state machines, and the DM, in order to implement the STDP rules on a traditional weight-changing neural topology.

## VI. Conclusion

Spiking neural networks with propagation delay can store information by trapping spikes within bistable loops. This kind of memory retain information immediately after stimulated. The memory model presented here neither uses synaptic weight changes nor depends on external learning algorithm.

Several bistable loops in parallel can form a memory cell, which can store as many bits as wanted. It is possible to arrange several cells in parallel, but it introduces the necessity of identifying a single operating cell for storing or retrieving information. We presented a 'decoder' for this purpose, which is a functional block that activates only one output at certain moment. It connects a single input to one out of many output paths. It also acts as a relay circuit that directs spikes from a single source to one specific data path among many possible.

Usually, memories have many cells with many bits each cell. It can be useful to combine all the bits in the memory cells in a single output. We presented a 'selector' for this purpose. Selector is a functional block that receives many spike sources but connects only a single source to a unique output at certain time.

These two functional blocks are important for creating automated *datapaths* for spike streams within spiking neuron networks with propagation delays.

We tested the memory with some spike streams presenting numbers associated to attributes. The memory is able to perform classification of prime/non-prime numbers, although this memory can hold any correlation and can memorize anything represented as spikes. The topology worked perfectly even when subjected to considerable levels of noise injections.

## Acknowledgment

The authors thanks the colleagues from the BINAC Research Group, at the Universidade Federal de Pernambuco, for relevant discussions on matters related to this research.

## References

[1] D. Purves, G. J. Augustine, D. Fitzpatrick, W. C. Hall, A.-S. Lamantia, J. O. McNamara, and M. S. Williams, *NEUROSCIENCE. 3rd. ed.* Sunderland, Massachusetts U.S.A.: Sinauer Associates, Inc., 2004.

[2] E. R. Kandel, J. H. Schwartz, and T. M. Jessel, *Principles of Neural Science. 4th ed.* New York, NY: McGrall-Hill Health Prof. Division, 2000.

[3] R. C. O'Reilly and Y. Munakata, *Computational Explorations in Cognitive Neuroscience: Understanding the Mind by Simulating the Brain*. 2000.

[4] J. Ranhel, "Neural Assembly Computing," *IEEE Trans. Neural Networks Learn. Syst.*, vol. 23, no. 6, pp. 916–927, Jun. 2012.

[5] [5] J. Ranhel, C. V Lima, J. Monteiro, J. E. Kogler Jr., and M. L. Netto, "Bistable Memory and Binary Counters in Spiking Neural Network," in *2011 IEEE Symposium on Foundations of Computational Intelligence. FOCI 2011 Proceedings*, 2011, vol. 1, pp. 66–73.

[6] J. Ranhel, "Neural Assemblies and Finite State Automata," in *Proceedings of 1st BRICS Countries Congress (BRICS-CCI) and 11th Brazilian Congress (CBIC) on Computational Intelligence*, 2013, vol. 1, pp. 1–6.

[7] J. R. Oliveira-Neto, F. D. Belfort, R. Cavalcanti-Neto, and J. Ranhel, "Magnitude Comparison in Analog Spiking Neural Assemblies," in *Proceedings of IEEE World Congress on Computational Intelligence(IEEE WCCI) and International Joint Conference on Neural Networks (IJCNN 2014)*, 2014, vol. 1, pp. 1–6.

[8] J. P. C. Cajueiro and J. Ranhel, "Limits of Coincidence Detector Neurons as Decoders of Polychronous Neuronal Groups Firing Completely," in *The International Joint Conference on Neural Networks, IEEE, IJCNN-2015*, 2015, no. 1, pp. 1–5.

[9] E. M. Izhikevich, "Simple model of spiking neurons," *IEEE Trans. Neural Networks*, vol. 14, no. 6, pp. 1569–1572, 2003.

[10] E. M. Izhikevich, "Which model to use for cortical spiking neurons?," *IEEE Trans. Neural Networks*, vol. 15, no. 5, pp. 1063–1070, 2004.

[11] C. Eliasmith and C. H. Anderson, *Neural Engineering: computation, Representation, and Dynamics in Neurobiological Systems*. Cambridge, MA: A Bradford Book - MIT Press, 2004.

[12] G. Buzsáki, "Neural Syntax: Cell Assemblies, Synapsembles, and Readers," *Neuron*, vol. 68, no. 3, pp. 362–385, 2010.

[13] E. M. Izhikevich, "Polychronization: computation with Spikes," *Neural Comput.*, vol. 18, no. 2, pp. 245–282, 2006.

[14] D. E. Feldman, "The Spike-Timing Dependence of Plasticity," *Neuron*, vol. 75, pp. 556–571, 2012.

[15] J. Sjöström and W. Gerstner, "Spike-timing dependent plasticity," *Scholarpedia*, vol. 5, no. 2, p. 1362, 2010.